\documentclass{article}
\usepackage{spconf,amsmath,graphicx}
\usepackage{enumitem}

\usepackage{url}

\usepackage{hyperref} 

\title{Explaining Deep Models through Forgettable Learning Dynamics}
\name{Ryan Benkert, Oluwaseun Joseph Aribido and Ghassan AlRegib}
\address{
School of Electrical and Computer Engineering\\
                            Georgia Institute of Technology,\\
                            Atlanta, GA,  30332-0250, USA \\
                            \{rbenkert3, oja, alregib\}@gatech.edu}
\begin{document}

\onecolumn 

\begin{description}[labelindent=-1cm,leftmargin=1cm,style=multiline]

\item[\textbf{Citation}]{R. Benkert, O.J. Aribido, and G. AlRegib, “Explaining Deep Models Through Forgettable Learning Dynamics,” in IEEE International Conference on Image Processing (ICIP), Anchorage, AK, Sep. 19-22 2021} \\


\item[\textbf{Review}]
{
Date of acceptance: June 2021
} \\


\item[\textbf{Bib}] {@ARTICLE\{benkert2021\_ICIP,\\ 
author=\{R. Benkert, O.J. Aribido, and G. AlRegib\},\\ 
journal=\{IEEE International Conference on Image Processing\},\\ 
title=\{Explaining Deep Models Through Forgettable Learning Dynamics\}, \\ 
year=\{2021\}\\ 
} \\


\item[\textbf{Copyright}]{\textcopyright 2022 IEEE. Personal use of this material is permitted. Permission from IEEE must be obtained for all other uses, in any current or future media, including reprinting/republishing this material for advertising or promotional purposes,
creating new collective works, for resale or redistribution to servers or lists, or reuse of any copyrighted component
of this work in other works. }
\\
\item[\textbf{Contact}]{\href{mailto:rbenkert3@gatech.edu}{rbenkert3@gatech.edu}  OR \href{mailto:alregib@gatech.edu}{alregib@gatech.edu}\\ \url{http://ghassanalregib.info/} \\ }
\end{description}

\thispagestyle{empty}
\newpage
\clearpage
\setcounter{page}{1}

\twocolumn
%
\maketitle
\begin{abstract}
Even though deep neural networks have shown tremendous success in countless applications, explaining model behaviour or predictions is an open research problem. In this paper, we address this issue by employing a simple yet effective method by analysing the learning dynamics of deep neural networks in semantic segmentation tasks. Specifically, we visualize the learning behaviour during training by tracking how often samples are learned and forgotten in subsequent training epochs. This further allows us to derive important information about the proximity to the class decision boundary and identify regions that pose a particular challenge to the model. Inspired by this phenomenon, we present a novel segmentation method that actively uses this information to alter the data representation within the model by increasing the variety of difficult regions. Finally, we show that our method consistently reduces the amount of regions that are forgotten frequently. We further evaluate our method in light of the segmentation performance.
\end{abstract}
\section{Introduction}
\label{sec:intro}

\begin{figure}[h]
\begin{center}
\includegraphics[scale=0.6]{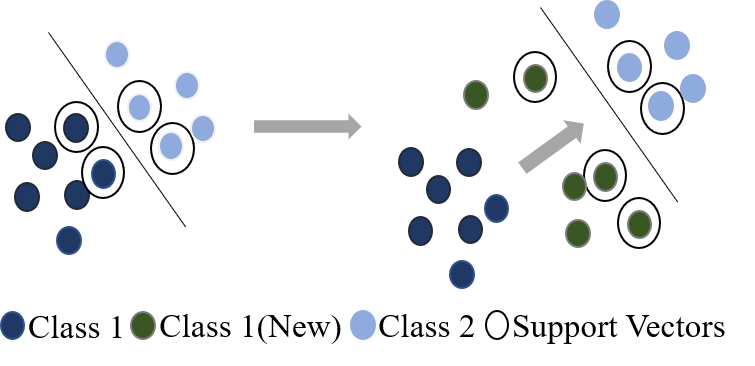}
\caption{Intuition of our support vector augmentation method}
\label{fig:sv-augmentation-intuition}
\end{center}
\end{figure}

Over the last decade, deep learning has had an impact on nearly every sector. It has paved the way for scientific breakthroughs in areas ranging from image recognition to complex medical diagnostics. The success of deep neural models lies in their ability to learn complex non-linear functions and estimate distributions of high dimensional data. In addition, open-source deep learning libraries enable fast large-scale deployment, making state-of-the-art algorithms available for countless applications. A central component of neural networks is how well they are capable of representing the target data. 
Well designed models can capture unique representations of the data and "learn" a function with a small error margin. In contrast, poor representations are often inconsistent and can produce semantically incorrect predictions. Therefore, understanding how the model represents and interacts with the data remains a very challenging and highly relevant research problem. One application, where this behaviour is especially important, is deep learning models for computational seismic interpretation. In seismic, there is limited open-source annotated data due to the high cost associated with data acquisition and expert annotation. For this reason, architectures designed for large computer vision applications over-fit on limited annotated seismic data and result in poor generalization capabilities. Due to the high relevance in this field, we present our method using the F3 block dataset (\cite{alaudah2019machine}) where several classes are underrepresented. Nevertheless, the work is applicable to a wide range of 2D data. \\
In this paper, we view neural networks in the context of their learning dynamics. Specifically, neural networks do not learn continually but forget samples over time. One branch of research investigates the forgotten information when a model is trained on one task but fine-tuned on another. In literature, this is often referred to as catastrophic forgetting (\cite{kirkpatrick2017overcoming, ritter2018online}). In contrast, \cite{toneva2018empirical} view the dynamics within a single data distribution and track the frequency in which information is forgotten during training. In this paper, we build upon this intuition and visualize frequently forgotten regions in a generalized segmentation framework. Similar to uncertainty works with Bayesian inference (\cite{kendall2017uncertainties}) or gradient based explanations (\cite{selvaraju2017grad, prabhushankar2020contrastive, lee2020gradients}), we can identify difficult regions and explain segmentation predictions. In contrast to other explainability techniques, frequently forgotten regions contain valuable information about the position within the representation space. Specifically, frequently forgotten regions are closer to the decision boundary and pose a threat to the generalization performance. Based on these findings we engineer a method that identifies challenging pixels and generates new samples that actively influence the representation mapping. In Fig.~\ref{fig:sv-augmentation-intuition} we show a toy example of our method. Based on the identified support vectors (circled blue disks), we generate new samples (green) that actively shift the decision boundary (black line) to reduce the amount of support vectors for a specific class. In contrast to traditional data augmentation (\cite{orr2003neural}), our method is data-driven and consistently reduces support vectors within the model.\\
The following are our contributions: First, we visualize difficult regions in the data by analyzing the learning dynamics during training. Second, we develop a augmentation method that reduces prone regions by actively shifting the decision boundary. Lastly, we compare our technique to popular augmentation techniques in literature.

\section{Methodology}
\label{sec:methodology}
Our goal is to quantify the learning behaviour during image segmentation by analysing the frequency of sample forgetting. In this section, we formally define when a sample is forgotten and how this relates to the proximity of samples to the decision boundary. Furthermore, we exploit these dynamics to actively shift the decision boundary in our model. Specifically, we identify support vectors in our training images and increase the variety through style transfer.
\subsection{Forgetting Events}
\label{sec:forgetting-events}
Intuitively, a sample is forgotten if it was classified correctly in a previous epoch and miss-classified in the current epoch. More formally, for image $I$ with (pixel, annotation) tuples ($x_i, y_i$), we define the accuracy of each pixel at a epoch $t$ as
\begin{equation}
acc^{t}_{i} = \mathbf{1}_{\tilde{y}^t_i = y_i}.
\end{equation}

\begin{figure}
\begin{center}
\includegraphics[scale=0.28]{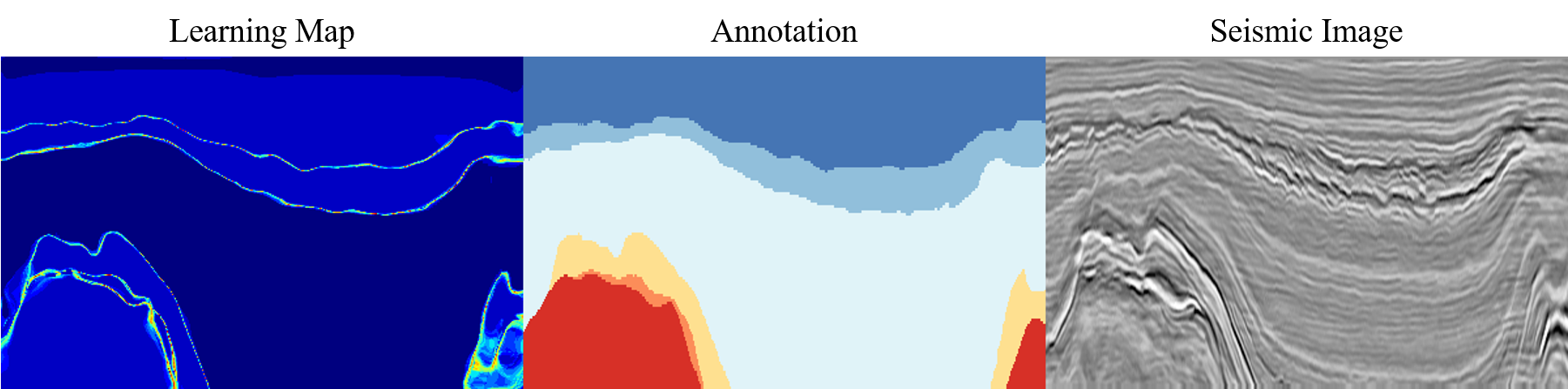}
\caption{An example of a forgetting event heat map as well as its corresponding image and annotation. Pixels close to the decision boundary are highlighted in different shades of red whereas pixels deep within the class manifold are dark blue. Note, that several classes (e.g the orange class "scruff") are underrepresented}
\label{fig:example-forgetting-example}
\end{center}
\end{figure}

Here, $\mathbf{1}_{\tilde{y}^t_i = y_i}$ refers to a binary variable indicating the correctness of the classified pixel in image $I$. With this definition we say a pixel was forgotten at epoch $t+1$ if the accuracy at $t+1$ is strictly smaller than the accuracy at epoch $t$:
\begin{equation}
f_{i}^{t} = int( acc^{t+1}_{i} < acc^{t}_{i} ) \in {1, 0}
\end{equation}
Following \cite{toneva2018empirical}, we define the binary event $f_i^t$ as a \emph{forgetting event} at epoch $t$. Since our application is a segmentation setting, we further visualize forgetting events in the spatial domain. Specifically, we count the amount of forgetting events occurring at each pixel $i$ and display them in a heat map. Mathematically, heat map $L \in \mathbf{N_{0+}}^{M\times N}$ is the sum over all forgetting events $f_i^t$ that occurred in time frame $T$:
\begin{equation}
L_{i} = \sum_{t = 0}^{T}{f_{i}^{t}}
\end{equation}
For better illustration, we present an example of a heat map in Fig~\ref{fig:example-forgetting-example}. Areas that were forgotten frequently are highlighted in shades of red in contrast to pixels that were forgotten rarely (blue). Similar to \cite{toneva2018empirical}, we can broadly classify the pixels into two groups: The first group consists of the pixels that were never forgotten or forgotten only rarely (e.g. light blue class in the center of Fig~\ref{fig:example-forgetting-example}). Since every epoch represents a model update, we conclude that these pixels are never or only rarely shifted outside of the class manifold in the feature space. In contrast, the second group consists of pixels forgotten more frequently (e.g. the class boundaries in Fig.~\ref{fig:example-forgetting-example}). Specifically, this means that several model updates shifted these pixels over the decision boundary during training, mapping them \emph{closer} to the decision boundary than unforgettable pixels. Similar to \cite{toneva2018empirical}, we argue that these pixels play a similar role to support vectors in maximal margin classifiers. In particular, we will show the importance of forgetting events in analyzing model predictions.
 
\subsection{Support Vector Augmentation}
\label{sec:sv-augmantation}

\begin{figure*}[h]
\begin{center}
\includegraphics[scale=0.55]{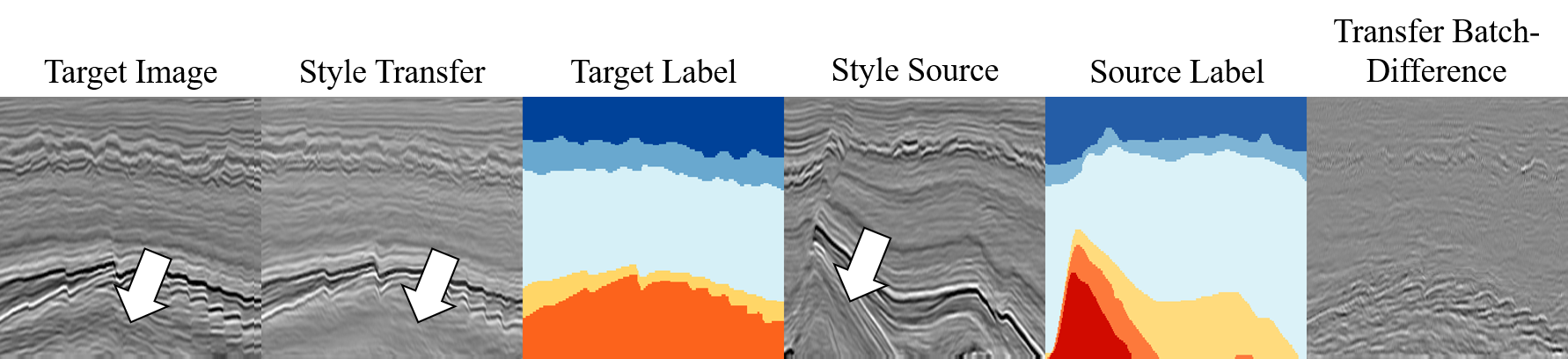}
\caption{Example of a feature transfer within two seismic images.}
\label{fig:subsurface-transfer}
\end{center}
\end{figure*}

\begin{figure}[h!]
\begin{center}
\includegraphics[scale=0.35]{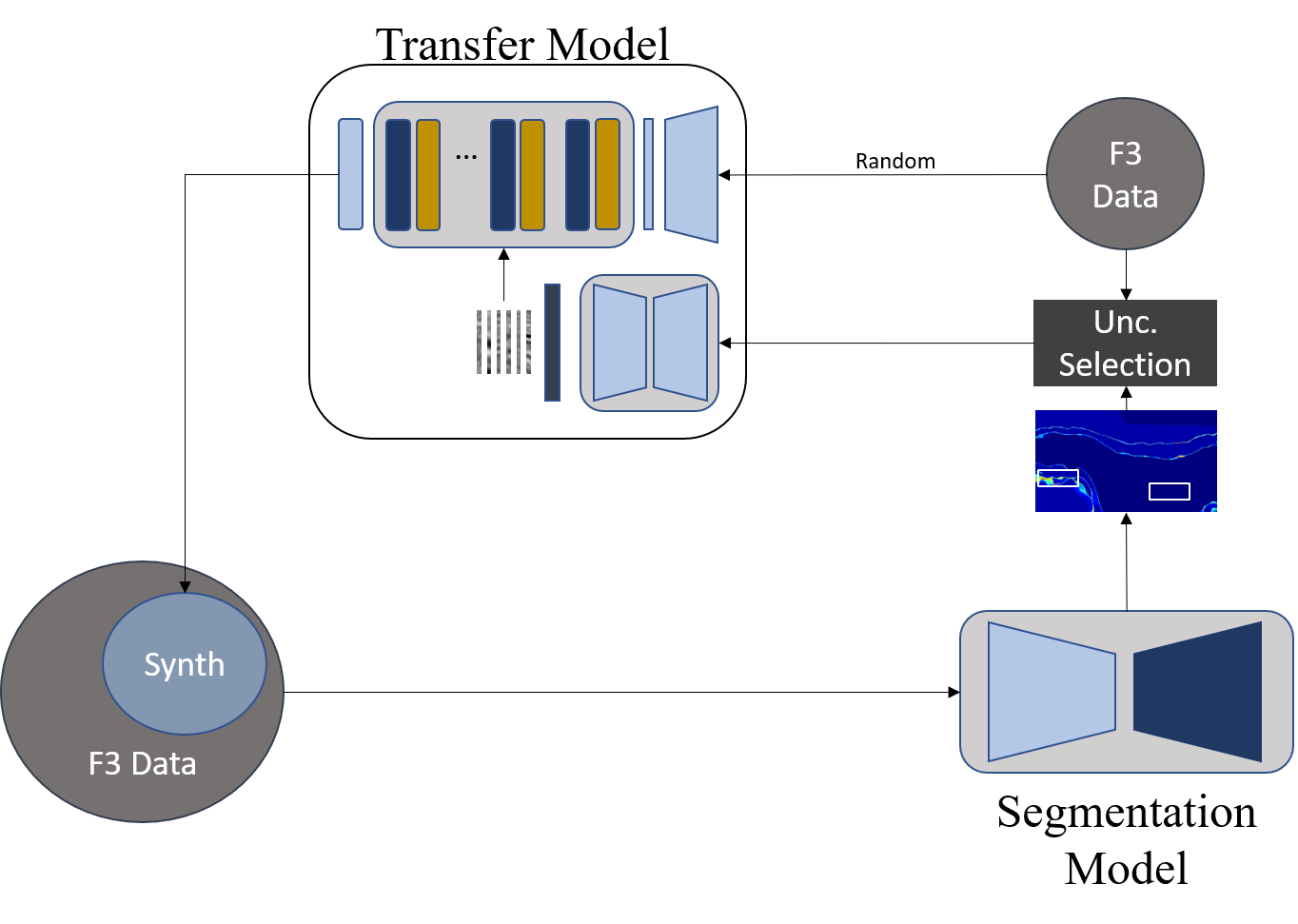}
\caption{Entire workflow of our architecture.}
\label{fig:subsurface-selection-architecture}
\end{center}
\end{figure}

As we have seen in Section~\ref{sec:forgetting-events}, forgetting events are a useful metric to quantify the sample position in the representation space. To be precise, forgetting events provide information about the proximity of samples to the decision boundary in a training interval $T$. In this section, we will exploit this information to increase the variety of forgettable pixels. In the seismic application, we achieve this through style transfer models (\cite{zhu2020sean, park2019semantic}). Specifically, we transfer class specific visual features from a source image to a target image without changing the structure or characteristics of neighboring classes. We target specific classes with a high forgetting event density and transfer the characteristics to other sections without affecting the geologic properties of the seismic images. An example of a style transfer is presented in Fig.~\ref{fig:subsurface-transfer}. Here, we show the target for the transfer, the resulting transfer image (second column from the left), the target annotation, and the style source with its corresponding label. The image on the far right of Fig.~\ref{fig:subsurface-transfer} shows the difference between the transfer images of subsequent batches with different style sources. In this example, we transfer the visual features of class "scruff" (orange) from the style source to the target image.  Moreover, switching the source image largely affects the target class (difference image in Fig.~\ref{fig:subsurface-transfer}) and presents the desired functionality of our algorithm.\\
Our method consists of a segmentation model, a transfer model and a data selection step (Fig~\ref{fig:subsurface-selection-architecture}). First, our method trains the segmentation model on the training data and produces a forgetting event heat map for every validation image in the training volume. In principle, heat maps could be produced for the entire training set but is computationally inefficient. In our implementation, the segmentation architecture is based on the deeplab-v3 architecture by \cite{chen2017deeplab} with a resnet-18 (\cite{he2016deep}) backbone. Our choice is based on empirical evaluations of performance and computational efficiency.\\
In the next step of our workflow, we calculate the forgetting event density within each class of a heat map. Specifically, we sum all forgetting events $f_{i\in c_k}$ within class $c_k$ of a heat map and divide by the number of pixels of class $c_k$ in the image. This metric allows us to rank each heat map according to its density with regard to an arbitrary class in the dataset.\\
Finally, we transfer the visual features of a predefined class from the images with the highest density to randomly sampled training images. Here, our architecture is a slightly altered version of \cite{zhu2020sean}. In short, the model modulates the style characteristics on the batch-normalization outputs within the image generator. This enables class specific transfers without affecting the geology of the image. In our method, we transfer the underrepresented classes within our data-set as these classes are generally most difficult to learn. After generation, the transferred images are added to our training pool and the segmentation model is trained from scratch.

\section{Results and Discussion}
\label{sec:results-and-discussion}

\begin{table*}[!h]
	\centering
	\begin{tabular}{ |p{2.3cm}||p{2.2cm}|p{2.2cm}|p{2.2cm}|p{2.2cm}|p{2.2cm}|p{2.2cm}|}
		\hline
		\multicolumn{7}{|c|}{Class Accuracy} \\
		\hline
		Class & Upper N. S. & Middle N. S. & Lower N. S. & Chalk & Scruff & Zechstein\\
		\hline
		Baseline & 0.982 & 0.912 & 0.969 & 0.816 & $\mathbf{0.383}$ & 0.651\\
		\hline
		Random Flip & 0.983 & 0.899 & 0.967 & 0.820 & $\mathbf{0.354}$ & 0.672\\
		\hline
		Random Rotate & 0.974 & 0.933 & 0.974 & 0.824 & $\mathbf{0.533}$ & 0.681\\
		\hline
		Ours & 0.982 & 0.906 & 0.966 & 0.810 & $\mathbf{0.438}$ & 0.656\\
		\hline
	\end{tabular}
	\caption{Averaged class accuracy over five augmentation experiments.}
	\label{table:results-overall}
\end{table*}

\begin{figure*}[!htb]
\begin{center}
\includegraphics[scale=0.6]{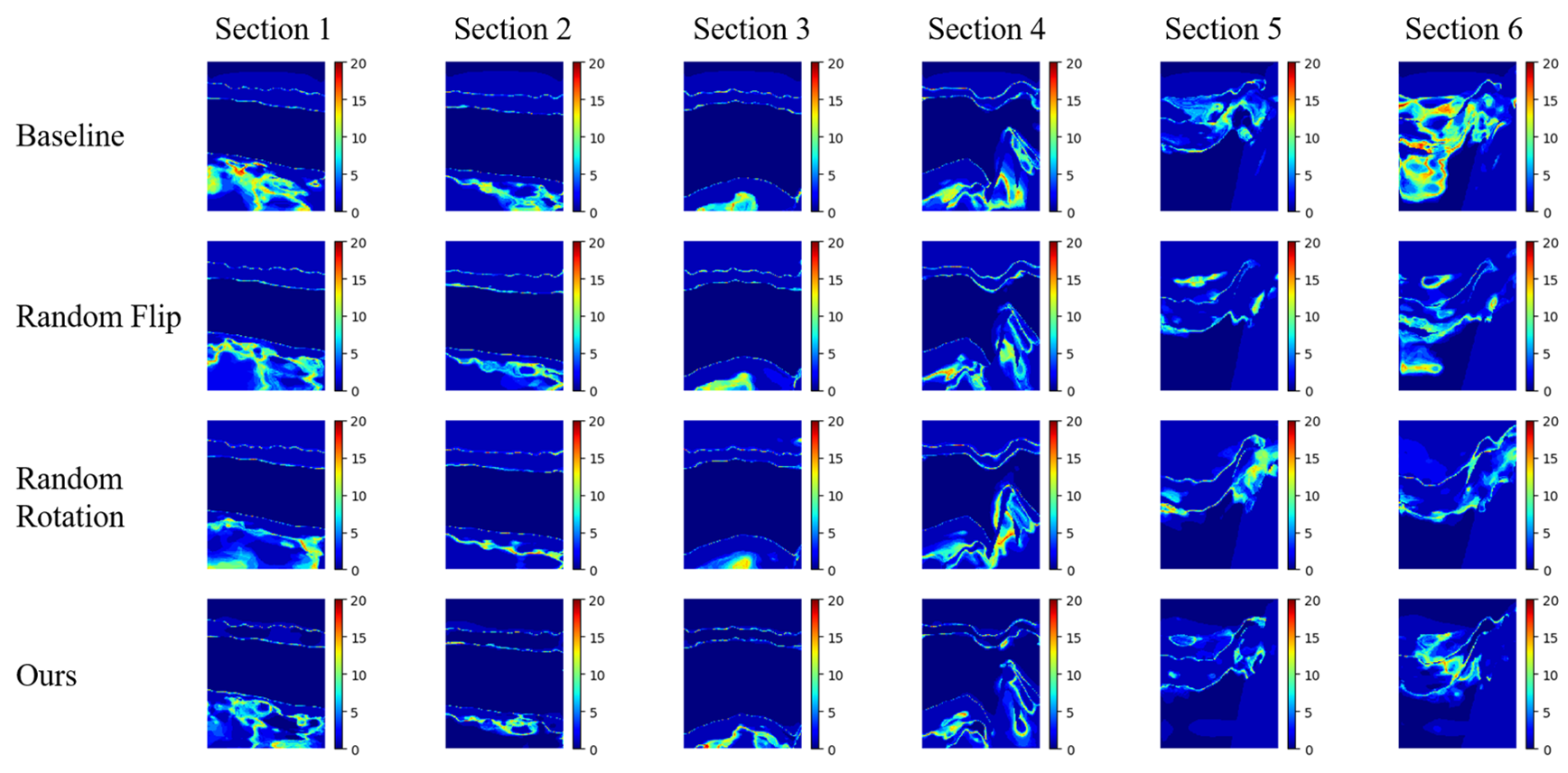}
\caption{Heat maps when using different augmentation methods. Our method significantly reduces the amount of forgetting events and impacts the regions shape.}
\label{fig:sv-augmentation}
\end{center}
\end{figure*}

To produce computationally efficient forgetting event heat maps, we train the network for 60 epochs and only track the validation and test set heat maps. In each of our experiments, the validation set is chosen by selecting every fifth vertical slice (referred to as inlines) and horizontal slice (referred to as crosslines) of the training volume. Subsequently, we query six images with the highest forgetting event density of our target class. Each image is used as a style source to generate 64 transfer images. For generation, we sample randomly to obtain the target image and retrain the segmentation model from scratch. In this paper, we only report the results when transferring the orange class (scruff). Other underrepresented classes (e.g. the red class zechstein) rendered similar results and are omitted. In our numerical analysis, our results are averaged over five separate experiments to account for random factors (e.g. initialization). We compare our method to other common augmentation methods (random horizontal flip and random rotations) in terms of segmentation performance (in class accuracy) and the forgetting event heat maps. The results are shown in Table~\ref{table:results-overall} and Fig.~\ref{fig:sv-augmentation} respectively.
Overall, our method reduces the amount of forgetting events significantly more than other augmentation methods. Specifically, we find that several regions with a high forgetting event density are transferred to a low density or disappear entirely (bottom class in Section~2 or entire right part of Section~6). These regions were shifted away from the decision boundary and model updates had little or no affect on the classification accuracy during training. In contrast, we find that no forgetting event regions disappear in the standard augmentation methods. Instead, the severity of forgetting event regions is reduced.\\
Numerically, all methods overwhelmingly match or outperform the baseline with respect to class accuracy. We note, that our method only affects the scruff class accuracy and matches the baseline performance of all other classes. This shows flexibility in our algorithm and allows an increased control over the network performance. We further observe that random rotations outperform our technique even in the scruff class. Although the class accuracy is higher, the forgetting event maps show significantly more forgetting event regions than the maps produced by our method. Moreover, the locations and shapes of the 
prone regions produced by the traditional methods are similar to the baseline regions (e.g. bottom class of Section~3). In contrast, our method changes the shape and location of the severe forgetting event region indicating a clear shift in the representation space.\\
Finally, we also identify regions with a lower forgetting event density that transitioned to a higher density (Section~5 bottom left) by applying our method. This allows us to analyze model weaknesses and interpret the segmentation output in light of training difficulty.
\section{Conclusion}
\label{sec:conclusion}
In this paper, we explain the behaviour of deep models by tracking how often samples are forgotten in between model updates. We identify regions that are especially difficult for the model and evaluate how these regions change when different segmentation strategies are pursued. Finally, we engineer a novel method that explicitly exploits this characteristic to actively influence how the data is represented within the model. We show that our method increases the margin of difficult regions indicating a clear decision boundary shift.

\vfill
\pagebreak


\bibliographystyle{ieeetr}
\bibliography{refs}

\end{document}